\documentclass[10pt,twocolumn,letterpaper]{article}
\usepackage{cvpr}
\usepackage{pdfpages}
\usepackage[accsupp]{axessibility}

\newcommand{\mparagraph}[1]{\vspace{0.5mm}\noindent{\textbf{#1.}\hspace{1mm}}}

\usepackage{import} %
\graphicspath{{figs/}}

\usepackage[ruled,noend]{algorithm2e} %

\SetAlFnt{\small}
\SetAlCapFnt{\small}
\SetAlCapNameFnt{\small}
\SetAlCapHSkip{0pt}
\IncMargin{-\parindent}
\usepackage{ifpdf}
\usepackage{graphicx}
\ifpdf
  
\else
\fi
\usepackage{color}
\usepackage{colortbl}
\definecolor{cyan}{cmyk}{1,0,0,0}
\definecolor{darkgreen}{rgb}{0,0.5,0}
\definecolor{orange}{rgb}{1,0.5,0}
\definecolor{magenta}{cmyk}{0,1,0,0}
\definecolor{darkyellow}{cmyk}{0,0,0.75,0}
\definecolor{gray}{rgb}{0.8,0.8,0.8}

\usepackage{amsmath} %

\usepackage[toc,page]{appendix} %
\usepackage{wrapfig} %
\usepackage{comment}
\usepackage{bm}
\usepackage{cuted} %
\usepackage{lipsum} %
\usepackage{mathtools} %
\usepackage{algorithmicx}
\usepackage{algpseudocode}
\usepackage{nicefrac}

\usepackage{gensymb}
\usepackage{float}
\usepackage{soul}
\usepackage{array}
\usepackage{multirow}
\usepackage{hhline}
\usepackage{setspace}
\usepackage{nicefrac}

\usepackage{diagbox}
\usepackage{makecell}
\usepackage{graphicx}
\usepackage{adjustbox}
\usepackage{caption}
\usepackage{booktabs}
\usepackage{tabularx}
\usepackage{subcaption} 
\algdef{SE}[DOWHILE]{Do}{doWhile}{\algorithmicdo}[1]{\algorithmicwhile\ #1}%
\usepackage{microtype}

\makeatletter
\renewcommand{\ALG@beginalgorithmic}{\small}
\makeatother

\newcommand{\DELETE}[1]{} %
\newcommand{\IGNORE}[1]{}
\usepackage{datenumber}
\usepackage{calc}
\usepackage[mmddyyyy]{datetime}

\newcounter{datetoday}
\newcounter{diffyears}
\newcounter{diffmonths}
\newcounter{diffdays}

\newcommand{\difftoday}[3]{%
      \setmydatenumber{datetoday}{\the\year}{\the\month}{\the\day}%
      \setmydatenumber{diffdays}{#1}{#2}{#3}%
      \addtocounter{diffdays}{-\thedatetoday}%
      \ifnum\value{diffdays}>0
        \def\diffbefore{}%
        \def\diffafter{left}%
      \else
        \def\diffbefore{}%
        \def\diffafter{ago}%
        \setcounter{diffdays}{-\value{diffdays}}%
      \fi
      \setcounter{diffyears}{\value{diffdays}/365}%
      \setcounter{diffdays}{\value{diffdays}-365*\value{diffyears}}%
      \setcounter{diffmonths}{\value{diffdays}/30}%
      \setcounter{diffdays}{\value{diffdays}-30*\value{diffmonths}}%
      \diffbefore
      \ifnum\value{diffyears}=0
      \else
        \ifnum\value{diffyears}>1
            \thediffyears\space years,
        \else
            \thediffyears\space year,
        \fi
      \fi
      \ifnum\value{diffmonths}=0
      \else
        \ifnum\value{diffmonths}>1
            \thediffmonths\space months
        \else
            \thediffmonths\space month
        \fi
      \fi
      \ifnum\value{diffdays}=0
      \else
        \ifnum\value{diffdays}>1
            \thediffdays\space days
        \else
            \thediffdays\space day
        \fi
      \fi
      \diffafter
}

\def\thickhline{\noalign{\hrule height 1pt}}

\definecolor{cvprblue}{rgb}{0.21,0.49,0.74}
\usepackage[pagebackref,breaklinks,colorlinks,allcolors=cvprblue]{hyperref}

\def\paperID{}
\def\confName{CVPR}
\def\confYear{2026}

\title{Dense Metric Depth Completion from Sparse Direct Time-of-Flight Sensors}

\author{Hakyeong Kim\footnotemark[1]\\
KAIST\\
{\tt\small hkkim@vclab.kaist.ac.kr}
\and
Ruicheng Wang\footnotemark[1]\\
USTC\\
{\tt\small t-ruiwang@microsoft.com}
\and
Chengtang Yao\\
Microsoft Research Asia\\
{\tt\small chengtangyao@microsoft.com}
\and
Jiaolong Yang\\
Microsoft Research Asia\\
{\tt\small jiaoyan@microsoft.com}
\and
Min H. Kim\\
KAIST\\
{\tt\small minhkim@vclab.kaist.ac.kr}
}

\begin{document}
\twocolumn[
\maketitle
{
\begin{center}
  \vspace{-0.75cm}
    \includegraphics[width=\linewidth]{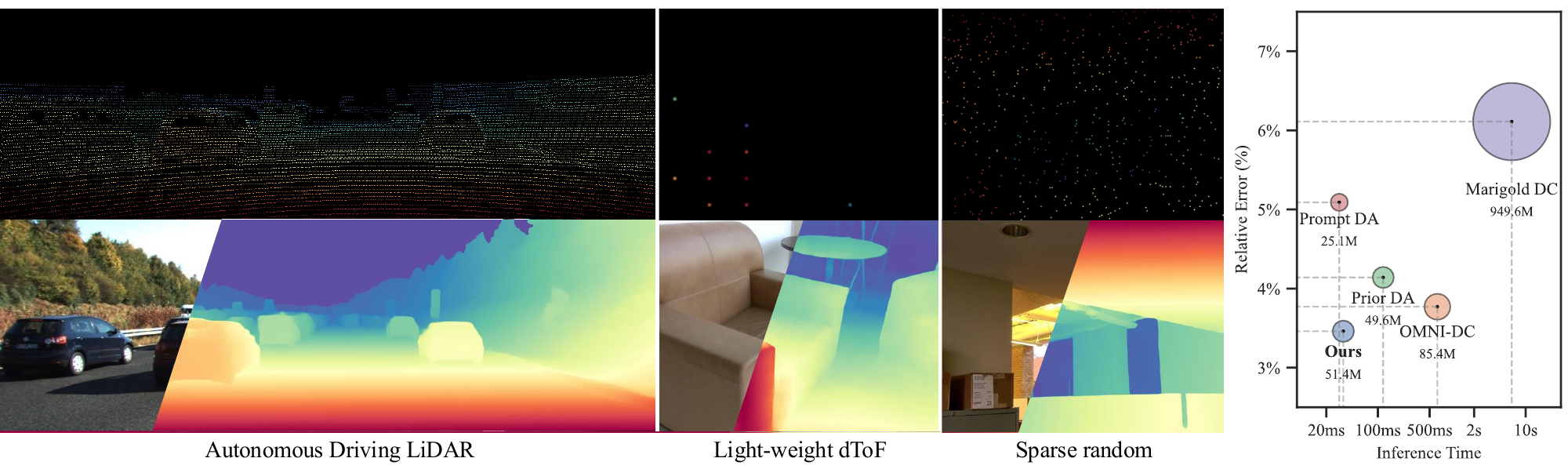}%
    \vspace{-3mm}
  \captionof{figure}{
  Zero-shot generalization of our model across different dToF sensing conditions.
Top: sparse depth inputs from three representative settings—autonomous driving LiDAR (rotating dToF), lightweight mobile dToF, and extremely sparse random sampling.
Bottom: our predicted dense metric depth for each case, demonstrating strong robustness under diverse sparsity, noise, and sensor patterns.
Right: comparison of accuracy versus inference time (bubble size indicates model parameters). Our method achieves the most favorable balance of low error and fast runtime, outperforming state-of-the-art depth completion and enhancement approaches. See Table~\ref{tab:depth_completion} for numerical details.}
  \label{fig:teaser}
\end{center}
}]

\footnotetext[1]{Work done during internship at Microsoft Research Asia.}
\begin{abstract}
Direct Time-of-Flight (dToF) sensors provide highly accurate metric depth and are more robust than indirect ToF systems in challenging real-world conditions. 
However, their high manufacturing cost and limited photodiode array size produce depth maps that are extremely sparse, low-resolution, and noisy, making them unsuitable for VR/XR, robotics, and 3D perception tasks that require dense metric depth. 
Existing monocular and depth completion methods struggle to handle the unique sampling patterns and hardware artifacts of dToF devices, and their performance often deteriorates significantly under severe sparsity or noise.
We present a generalizable framework for dense metric depth completion from sparse dToF measurements, capable of operating across diverse sensor types, sparsity levels, and noise conditions. 
Our model employs a depth-guided dual-branch Vision Transformer encoder that processes RGB images and sparse dToF measurements separately, while a masked joint attention module allows depth tokens to reliably guide image features without being overwritten by them. 
A lightweight decoder reconstructs dense metric depth efficiently, without diffusion-based or refinement-heavy post-processing.
To address the scarcity of paired training data, we introduce a comprehensive dToF simulation pipeline that reproduces the characteristics of flash, sub-VGA flash, and rotating sensors, including hardware-induced degradation, irregular sparsity, and realistic noise distributions. 
Trained entirely on synthetic data, our model achieves strong zero-shot generalization across 6 datasets and 3 real dToF devices, outperforming state-of-the-art approaches in both accuracy and computational efficiency. 
This establishes a robust and practical solution for dense metric depth completion from sparse direct ToF sensors.
Our code and models will be open-sourced. See \href{https://vclab.kaist.ac.kr/cvpr2026p3}{project page}.
\end{abstract}

\section{Introduction}
\label{sec:intro}
Dense and high-quality depth perception is essential for VR/XR systems, robotics, and general 3D scene understanding. 
Although recent advancements in monocular depth estimation have shown impressive in-the-wild performance by training large-scale foundation models \cite{yang2024depthv2,wang2025moge,ke2024repurposing,hu2024metric3d,wang2025moge2}, these approaches still suffer from scale ambiguity and often produce inconsistent metric depth in complex real-world scenarios. 
This limitation makes them unreliable for applications that require precise and stable metric depth.

A common strategy to mitigate scale ambiguity is to incorporate sparse direct Time-of-Flight (dToF) measurements, such as LiDAR or low-resolution ToF depth, which provide accurate range information. 
While this integration has significantly improved robustness in various real-world settings \cite{park2024depth,lin2025prompting,liu2024depthlab}, existing methods are typically tailored to a specific depth sensor and struggle when the measurements become extremely sparse, noisy, or irregularly sampled. 
Furthermore, recent attempts that rely on diffusion models \cite{viola2025marigold}, iterative optimization \cite{zuo2025omni}, or multi-stage refinement \cite{wang2025depth} achieve strong performance but incur substantial computational overhead, limiting their practicality in mobile or real-time applications.

In this paper, we introduce a single model that generalizes effectively to diverse sparse dToF devices and challenging real-world scenarios, even under extreme sparsity or high noise, while remaining computationally efficient. 
We observe that many prior methods rely heavily on pretrained monocular encoders and treat sparse depth merely as an auxiliary signal, failing to capture the complementary and mutually constraining structure shared between RGB features and dToF measurements. 
To address this, we propose a depth-guided dual-branch encoder that independently processes RGB images and sparse dToF depth, while a masked joint attention module enables controlled information flow from depth tokens to RGB tokens without corrupting the depth representation. 
This design yields a more expressive and depth-aware feature space, allowing a lightweight decoder to produce high-quality dense metric depth without requiring complex refinement networks.

To further enhance generalization, we introduce a comprehensive dToF simulation pipeline that models a wide range of sensor characteristics, including flash, and rotating devices, along with realistic noise patterns, occlusion-induced missing regions, and hardware-level degradation. 
This scales training to large synthetic datasets and learns a robust representation that transfers well to real-world sparse dToF inputs.

We conduct extensive experiments across 6 datasets, 3 dToF sensor devices, and multiple sparsity and noise levels, covering indoor scenes and street environments.
Our pretrained model exhibits strong zero-shot generalization across all settings and outperforms state-of-the-art depth completion and fusion-based approaches in both accuracy and efficiency. 
Ablation studies further validate the effectiveness of our encoder architecture.

Our contributions are summarized as follows:
\begin{itemize}
\item We introduce a new framework for dense metric depth completion from sparse direct ToF measurements, achieving strong zero-shot generalization across diverse sensor devices and depth-related tasks using a single model trained solely on synthetic data.
\item We propose a depth-guided dual-branch encoder with masked joint attention, enabling effective and controlled feature exchange between RGB and sparse depth. This expressive encoder allows us to use a lightweight decoder without complex refinement while maintaining high accuracy and efficiency.
\item We design a comprehensive dToF simulation pipeline that reproduces realistic sensor noise, sparsity patterns, and hardware artifacts across multiple dToF device families, significantly improving synthetic-to-real transfer.
\end{itemize}

\section{Related Work}
\subsection{Monocular Depth Estimation}

Monocular depth estimation aims to infer 3D scene geometry from a single RGB image. Recent advances have demonstrated strong generalization in the wild by scaling up depth foundation models \cite{yang2024depth,yang2024depthv2,wang2025moge} or by leveraging diffusion-based generative priors \cite{ke2024repurposing,fu2024geowizard,helotus}. 
Although these approaches achieve high-quality relative depth, they are typically trained with scale-invariant losses and therefore cannot reliably recover absolute metric scale, which is required in many downstream applications.

To address scale ambiguity, several works estimate metric depth by direct scale regression from image features \cite{bochkovskiydepth,piccinelli2024unidepth,wang2025moge2}, 
while others perform scale alignment using camera intrinsics or geometric constraints \cite{yin2023metric3d,hu2024metric3d,guizilini2023towards}. 
Despite the more explicit treatment of scale, these methods still struggle with metric accuracy under challenging lighting, textureless regions, and complex outdoor conditions.

In contrast to monocular-only methods, we incorporate sparse dToF measurements as reliable geometric references. This allows our model to preserve absolute scale and produce high-quality dense metric depth across diverse real-world scenarios.

\begin{figure*}[ht]
	\centering
	\vspace{-3mm}
	\adjustbox{}{
		\includegraphics[width=0.8\linewidth]{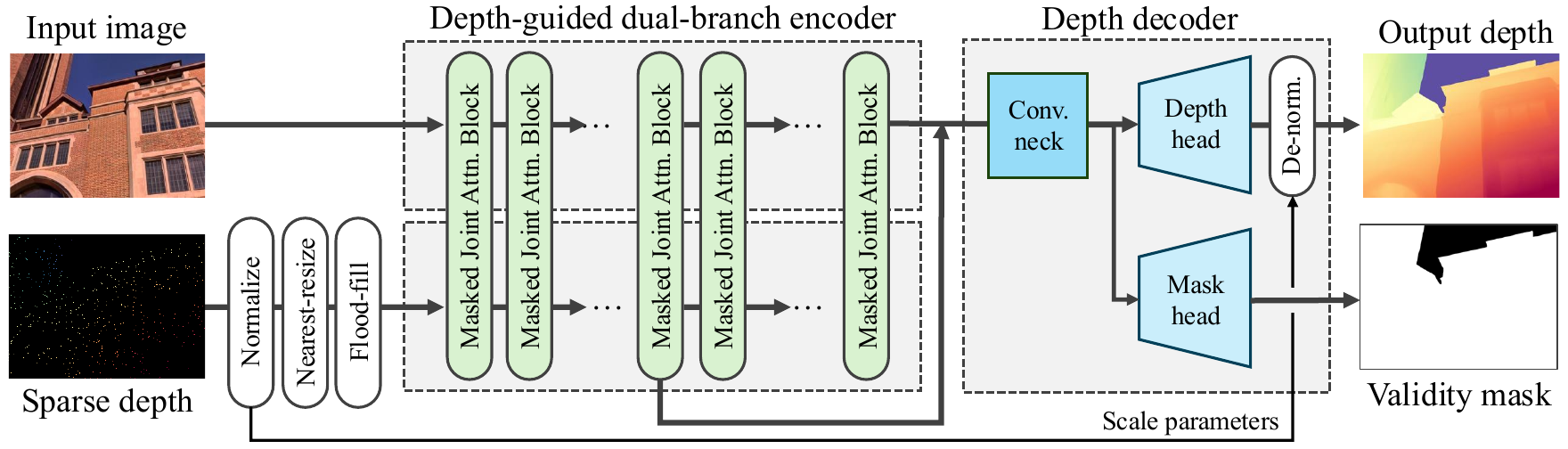}
	}
	\vspace{-3mm}
	\caption{\label{fig:method_overview}
	Overview of our method.
	Given an input RGB image and sparse dToF depth, we first normalize and upsample the sparse measurements to obtain a unified depth representation. 
	The RGB image and sparse depth are then encoded by a dual-branch ViT encoder, where multiple masked joint attention blocks enable depth tokens to guide RGB features without corrupting the depth representation. 
	The fused features are passed to a lightweight DPT decoder, which predicts dense normalized depth and a validity mask. 
	The final metric depth is recovered through de-normalization using scale parameters from preprocessing. Our design produces high-quality dense metric depth from sparse dToF inputs while remaining computationally efficient.}
	\vspace{-4mm}
\end{figure*}

\subsection{Depth Completion}
Depth completion seeks to infer dense depth from sparse measurements. 
Early methods densify sparse inputs by propagating depth values using hand-crafted priors or convolutional networks \cite{hu2021penet,tang2024bilateral,Wang_2023_ICCV,liu2017learning,cheng2019learning}. 
While effective, these models are usually designed for specific sensors or datasets, limiting generalization when applied to new environments or different sparsity patterns.

Recent studies aim for broader generalization by combining sparse depth with powerful monocular priors. 
PromptDA \cite{lin2025prompting} and DepthPrompting \cite{park2024depth} integrate monocular depth foundation models into completion pipelines, achieving strong zero-shot performance but remaining tied to specific sensor configurations. 
Diffusion-based approaches such as Marigold-DC \cite{viola2025marigold} and DepthLab \cite{liu2024depthlab} further improve generalization without task-specific training, yet they are sensitive to noise and extremely sparse inputs, and often require heavy computation.

Optimization-based frameworks such as OGNI-DC \cite{zuo2024ogni} and OMNI-DC \cite{zuo2025omni} perform iterative refinement at inference time, offering robustness but with significant runtime overhead. 
PriorDA \cite{wang2025depth} improves prediction quality by applying monocular priors in both coarse alignment and fine refinement stages, but this two-stage pipeline is computationally expensive.

Unlike the above methods, our approach does not rely on diffusion models, iterative optimization, or multi-stage refinement. 
Instead, we demonstrate that a highly efficient dual-branch Transformer encoder, combined with realistic dToF simulation, can achieve strong zero-shot generalization across diverse dToF devices, noise conditions, and sparsity levels while maintaining fast inference and practical applicability.

\begin{figure*}[t]
    \centering
    \vspace{-0.5cm}  
    \begin{minipage}[t]{0.48\textwidth}
        \centering
        \vspace{0pt}
        \includegraphics[width=\linewidth]{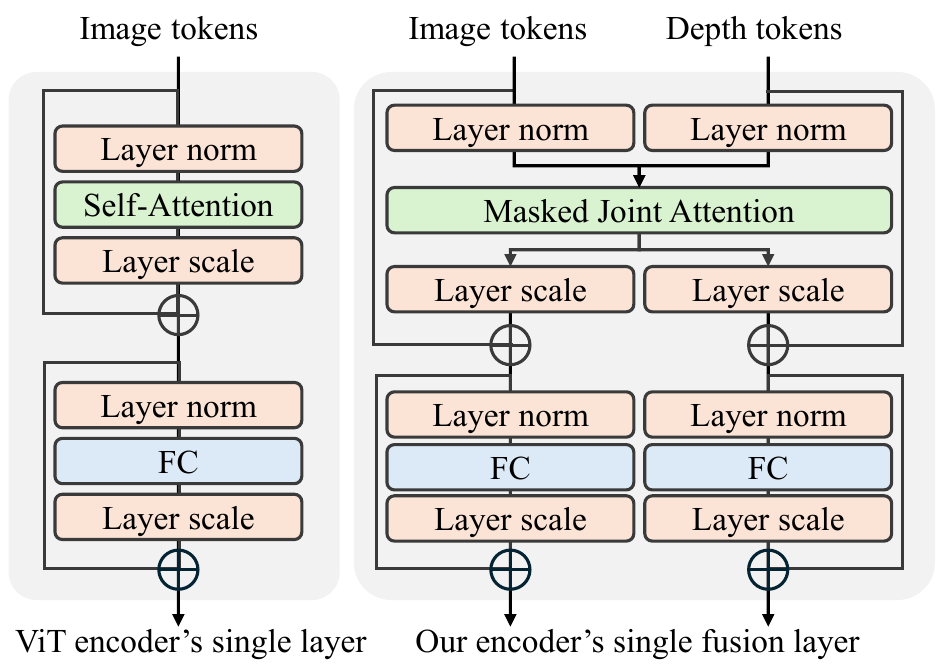}%
        \vspace{-0.2cm}        
        \caption{\label{fig:network_block}
Left: a single layer of the standard ViT encoder, which performs self-attention and feed-forward operations on image tokens only.
Right: our proposed fusion layer, where image and depth tokens are processed in two parallel branches and fused through a masked joint attention module. This design preserves the original ViT structure while enabling depth-guided feature interaction between the two modalities.}
        \vspace{-0.3cm}
    \end{minipage}
    \hfill
    \begin{minipage}[t]{0.48\textwidth}
        \centering
        \vspace{0pt}
        \includegraphics[width=\linewidth]{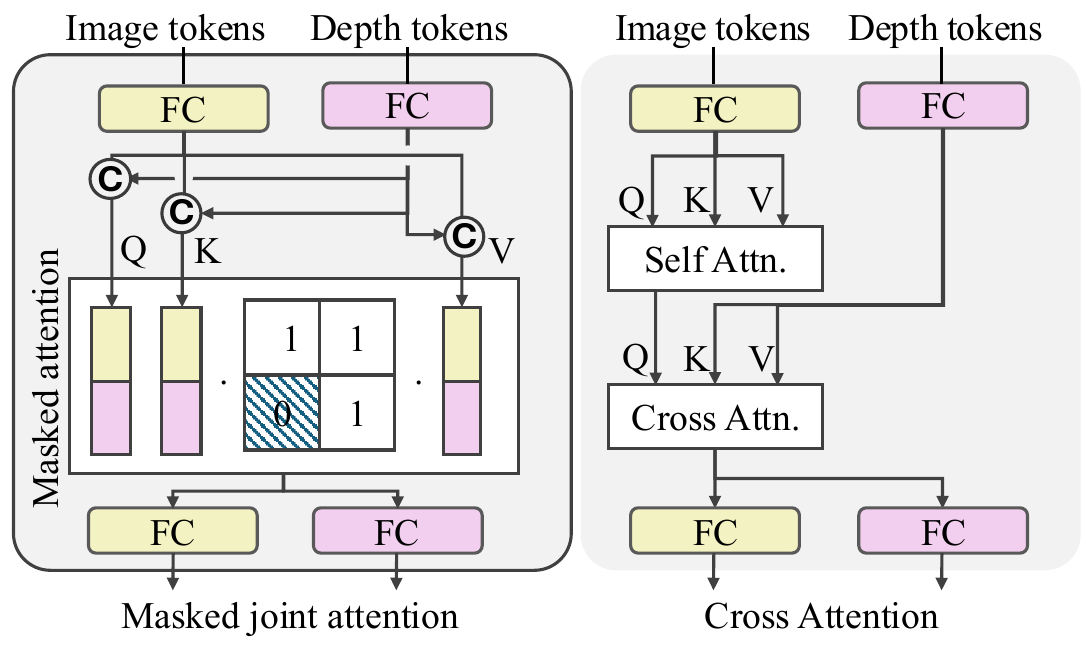}
        \caption{\label{fig:masked_cross_attention}
Left: in our masked joint attention, queries, keys, and values are computed jointly from the concatenated image and depth tokens. A directional mask is applied to restrict information flow from image tokens to depth tokens, while still allowing depth-to-image guidance. This enables effective and controlled fusion of spatial and geometric cues.
Right: in cross-attention, queries are taken from one modality while keys and values come from the other, leading to asymmetric dependence and modality-specific attention flows.}
        
    \end{minipage}
    \vspace{-0.5cm}
\end{figure*}

\section{Method}
\label{sec:method}
Given a high-resolution RGB image $\boldsymbol{I} \in \mathbb{R}^{H \times W \times 3}$ and a sparse depth map $\boldsymbol{Z} \in \mathbb{R}^{H \times W}$ captured by a dToF sensor, our goal is to estimate a dense, high-resolution metric depth map $\tilde{\boldsymbol{D}} \in \mathbb{R}^{H \times W}$ along with a validity mask $\tilde{\boldsymbol{M}} \in \mathbb{R}^{H \times W}$ indicating unreliable or unobservable regions (e.g., sky), as illustrated in Figure~\ref{fig:method_overview}. 
To accommodate diverse dToF sensor devices, we first preprocess the sparse depth input to construct a unified depth representation. 
We then employ a depth-guided dual-branch encoder that separately encodes RGB and sparse depth features while enabling controlled cross-modal fusion through a masked joint attention mechanism. 
This produces depth-aware image features that are subsequently decoded by a lightweight DPT-based decoder to generate dense metric depth. 
Lastly, we introduce a comprehensive synthetic data generation pipeline that models diverse dToF sensor characteristics, noise distributions, and sparsity patterns, significantly improving the model’s robustness and zero-shot generalization.

\subsection{Preprocessing}
\label{sec:preprocessing}
\mparagraph{Depth projection and filling}
To address the heterogeneous resolutions and sampling patterns of different dToF devices, we first convert the raw sensor depth into a unified representation.
Because dToF depth maps typically have much lower spatial resolution than the corresponding RGB image, we upsample both the depth map and its validity mask to the RGB resolution. 
We then fill missing-depth regions using nearest-neighbor interpolation to obtain a continuous depth field that preserves local geometric structures. 
Importantly, the validity mask remains sparse, allowing the network to distinguish between pixels originating from actual sensor measurements and those generated through interpolation.

\mparagraph{Depth normalization}
To enable the depth branch to use the pretrained ViT encoder from DINOv2~\cite{oquab2023dinov2}, we convert the sparse depth map into a three-channel tensor whose statistical distribution matches that of DINOv2's RGB input.
We first apply log-normalization to the sensor depth $\boldsymbol{Z}$ to stabilize scale variations and compress large depth ranges:
\begin{equation}
    \begin{aligned}
        \alpha &= \log (Z_{\max}) - \log (Z_{\min}),\\
        \beta  &= \log (Z_{\min}),\\
        \hat{\boldsymbol{Z}} &= (\log \boldsymbol{Z} - \beta) / \alpha,
    \end{aligned}
    \label{eq:depth_normalization}
\end{equation}
where $Z_{\min}$ and $Z_{\max}$ are the raw minimum and maximum valid depth measurements, respectively.
In practice, this normalization remains stable under noisy and outlier-heavy depth simulations.

We then form a three-channel depth input by duplicating the normalized depth for the first two channels and using the sparse validity mask as the third channel. 
This provides the encoder with both geometric intensity information and measurement reliability. The final tensor is linearly scaled to the range $[-1, 1]$ to align with the dynamic range expected by the pretrained ViT.

\subsection{Depth-Guided Dual-Branch Encoder}
\label{sec:architecture}
Most depth completion approaches rely on an off-the-shelf monocular encoder and treat sparse depth as an auxiliary input, overlooking the strong correspondence between RGB appearance and geometric cues. 
This often limits their robustness when the sparse depth becomes extremely irregular or noisy. 
In contrast, we design an encoder that explicitly models cross-modal interactions between RGB images and sparse dToF measurements.

Our encoder consists of two parallel Vision Transformer (ViT) branches~\cite{ranftl2021vision}, one for the RGB image and one for the normalized sparse depth input. 
Instead of processing the two modalities independently or relying on standard cross-attention, we adopt a masked joint attention mechanism that enables controlled information exchange between the branches (Figure~\ref{fig:network_block}).

In masked joint attention, tokens from the image $\left[Q_I, K_I, V_I\right]$ and depth branches $\left[Q_Z, K_Z, V_Z\right]$ are first concatenated, after which the queries, keys, and values are computed jointly:
\vspace{-1mm}
\[
Q=\begin{bmatrix}Q_I \\ Q_Z\end{bmatrix},\quad
K=\begin{bmatrix}K_I \\ K_Z\end{bmatrix},\quad
V=\begin{bmatrix}V_I \\ V_Z\end{bmatrix}.
\]
This differs from conventional cross-attention, where queries and keys/values come from different modalities (Figure~\ref{fig:masked_cross_attention}). We then apply a directional mask during the attention computation:
\begin{equation}
	\resizebox{0.9\linewidth}{!}{
	\mbox{\fontsize{10}{12}\selectfont $
    \text{Attention}(Q,K,V)
    = \text{softmax}\!\left(
        \frac{QK^{\top} \odot G}{\sqrt{d_k}}
    \right)V,
    \hspace{1mm} G=
\begin{bmatrix}
1 & 1 \\
0 & 1
\end{bmatrix}
$ } } %
\end{equation}

The mask $G$ enforces an asymmetric information flow: depth-to-image attention is allowed, enabling depth measurements to guide and refine image features, 
while image-to-depth attention is suppressed, preventing unreliable RGB cues from corrupting the sparse depth representation. 
This yields depth-aware image embeddings while preserving the integrity of the depth features.

An additional advantage is that masked joint attention closely matches the structure of the original ViT self-attention block. 
As a result, the overall encoder remains equivalent to two isomorphic ViT branches, allowing us to initialize both with pretrained DINOv2 weights~\cite{oquab2023dinov2}. 
This initialization injects strong visual and geometric priors from large-scale pretraining, significantly improving both convergence and generalization.

\subsection{Depth Decoder}
\label{sec:depth_decoder}
Our depth decoder follows the DPT architecture \cite{ranftl2021vision, wang2025moge} and consists of two lightweight decoding branches. The first branch predicts a single-channel validity mask, while the second branch estimates a dense normalized depth map $\hat{\boldsymbol{D}}$. To recover metric depth, we apply the same de-normalization parameters ${\alpha, \beta}$ used during preprocessing (Eq.~\eqref{eq:depth_normalization}):
\begin{equation}
    \tilde{\boldsymbol{D}} = \exp(\alpha \cdot \hat{\boldsymbol{D}} + \beta).
\end{equation}
Because $\hat{\boldsymbol{D}}$ is not constrained to the interval $[0,1]$, this formulation allows the decoder to extrapolate depth values beyond the original sensor range $[Z_{\min}, Z_{\max}]$. As a result, the model can infer plausible metric depth even in regions where the dToF sensor provides no reliable measurements.

\subsection{Loss}
\label{sec:loss}
We supervise our model using four complementary loss terms: depth-weighted L1 loss $\mathcal{L}{1}$, global scale-invariant loss $\mathcal{L}{g}$, local scale-invariant loss $\mathcal{L}{l}$, and a mask loss $\mathcal{L}{m}$:
\begin{equation}
\begin{aligned}
\mathcal{L}
= \mathcal{L}_{1} + \mathcal{L}_{g} + \mathcal{L}_{l} + \mathcal{L}_{m}.
\end{aligned}
\label{eq:example}
\end{equation}
We use equal weighting for all loss terms, which is consistent with prior work's~\cite{wang2025moge} design choice.

\mparagraph{Depth-weighted L1 loss}
We penalize the metric depth error between the predicted depth $\tilde{\boldsymbol{D}}=\{\tilde{d}_i\}$ and the ground truth $\boldsymbol{D}=\{d_i\}$ over all valid ground-truth pixels $\mathcal{M}$:
\vspace{-1mm}
\begin{equation}
\mathcal{L}_{1} = \sum_{i\in \mathcal M}\frac{1}{{d}_i} | \tilde{d}_i - {d}_i |_1,
\label{eq:loss_l1}
\vspace{-1mm}
\end{equation}
The inverse-depth weighting prevents the model from overfitting to large far-range values and encourages accurate estimation in near-range regions where geometric structures are more detailed.

\mparagraph{Global and local scale-invariant losses}
While L1 loss ensures metric correctness, it is insensitive to geometric shape, which may lead to distortions in object structures. To preserve both global and local geometry, we compute scale-invariant losses in 3D space.

We first project the predicted and ground truth depth maps into 3D point clouds, $\tilde{\boldsymbol{P}}=\{\tilde{\boldsymbol{p}}_i\}$ and ${\boldsymbol{P}}=\{\boldsymbol{p_i}\}$, using camera intrinsics. We then estimate a global scale factor $s$ using the ROE solver~\cite{wang2025moge}. For local geometry, we divide the image into patches $\{\mathcal{S}_j\}_{j\in \mathcal H}$, where $\mathcal{H}$ is the set of sampled anchor points, and compute patch-specific scales ${s_j}$.
The corresponding losses are:
\begin{equation}
\mathcal{L}_{g} = \sum_{i\in \mathcal M}\frac{1}{d_i} \|s \tilde{\boldsymbol{p}}_i - {\boldsymbol{p}}_i \|_1, \hspace{2mm}
\label{eq:loss_global}
\mathcal{L}_{l} = \sum_{j\in \mathcal H}\sum_{i\in\mathcal S_j} \frac{1}{d_i} \| s_j \tilde{\boldsymbol{p}}_i - {\boldsymbol{p}}_i \|_1.
\end{equation}
Together, these losses encourage the model to maintain a consistent metric scale while preserving fine-grained geometric structure.

\mparagraph{Mask loss}
Following MoGe~\cite{wang2025moge}, we also predict a validity mask $\tilde{\boldsymbol{M}}=\{\tilde{m}_i\}$ to identify regions with undefined or infinite depth (e.g., sky, reflective areas). The mask head is trained using binary cross-entropy:
\begin{equation}
\mathcal{L}_{m} = 
- \sum_{i}\left[
m_i \log(\tilde{m}_i) +
(1 - m_i) \log(1 - \tilde{m}_i)
\right],
\label{eq:loss_mask}
\vspace{-3mm}
\end{equation}
where $\boldsymbol{M}=\{m_i\}$ denotes the ground truth mask.
This term ensures reliable uncertainty handling in regions where the sensor cannot provide depth measurements.

\begin{figure}[pt]
	\centering
	\vspace{-2mm}
	\adjustbox{max width=0.8\linewidth}{
		\includegraphics[width=2\linewidth]{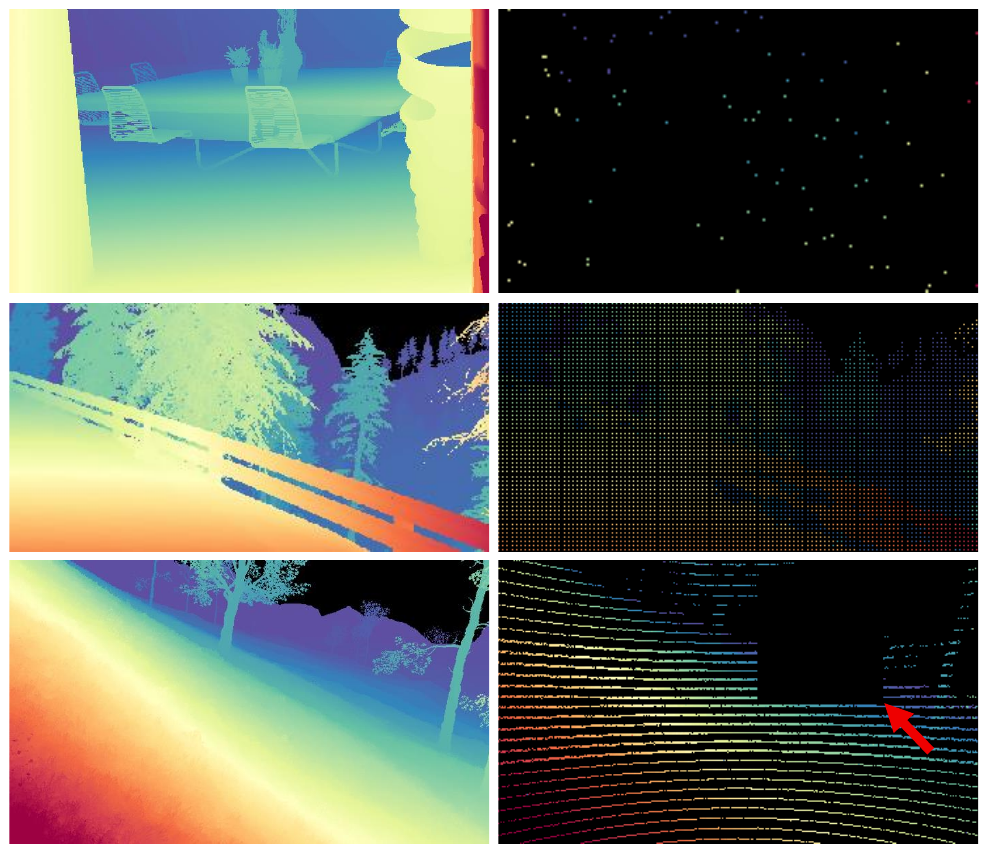}
	}
	\vspace{-0.3cm}
	\caption{\label{fig:data_simulation}
We simulate three representative patterns of direct ToF sensors: (top) extremely sparse Flash dToF, (middle) sub-VGA-resolution Flash dToF with image space projections, and (bottom) rotating LiDAR-style dToF with line-pattern sampling. The simulation reproduces realistic sparsity, resolution limits, and noise characteristics observed in real devices. The red arrow highlights an example of our depth inpainting augmentation, which models occlusions and missing returns %
encountered in real-world sensing.}

    \vspace{-7mm}
\end{figure}

\subsection{Sparse dToF Depth Simulation}
\label{label:training_details}
High-resolution ground-truth depth paired with real dToF sensor measurements is extremely scarce and limited in diversity. 
To overcome this constraint, we adopt a simple yet powerful strategy: we synthesize sparse dToF depth from a wide range of large-scale synthetic RGB-D datasets \cite{deitke2023objaverse, guo2020gen, le21wacv, wang2020flow, roberts:2021, wang2019irs, niklaus20193d, li2023matrixcity, fonder2019mid, huang2018deepmvs, zheng2020structured3d, wrenninge2018synscapes, ros2016synthia, gomez2025all, Tosi2021CVPR, wang2020tartanair, ranftl2021vision}.
This simulation pipeline enables us to scale training to millions of diverse scenes, effectively compensating for the lack of real sensor data and significantly improving robustness and zero-shot generalization.

Direct ToF sensors vary widely in their scanning mechanisms, which in turn determine the sparsity pattern, resolution, and noise characteristics of their depth outputs. 
We categorize these devices into two representative families: flash dToF, and rotating dToF. %
Each type exhibits distinct sampling distributions—from extremely sparse dot patterns to low-resolution dense grids and line-structured scans. 
To ensure compatibility with real devices, we simulate each family using sensor-specific rules that reproduce their unique sparsity, non-uniform coverage, and noise behavior (Figure~\ref{fig:data_simulation}).
This diverse sensor-aware simulation plays a crucial role in teaching our model to handle the highly irregular and device-dependent patterns encountered in real-world dToF measurements.

\mparagraph{Flash dToF camera}
Flash dToF sensors, widely used in mobile devices, illuminate the entire scene at once and return depth values determined by the resolution of their SPAD array. 
This results in extremely sparse depth samples. 
To simulate this sensing pattern, we randomly sample 64–10K points from the ground-truth depth map, following real sensor specifications and recent depth completion settings~\cite{liu2024depthlab, viola2025marigold}.

In addition to sparse flash sensors, many commercial devices output sub-VGA–resolution dense depth maps produced through onboard post-processing or by embedding small SPAD arrays in larger layouts~\cite{baruch2021arkitscenes, yeshwanth2023scannet++}. 
Although denser, these maps exhibit characteristic edge degradation, spatial smoothing, and hardware-induced artifacts. To model this second class of flash dToF sensors, we first downsample the ground-truth depth using nearest-neighbor interpolation to preserve thin structures. 
We then apply random erosion and dilation to depth boundaries to reproduce the blurry and softened edges commonly observed in real sensor output. To introduce spatially coherent variations, we generate Perlin noise at the image resolution and use it to assign a smoothly varying erosion–dilation radius to each pixel.

\mparagraph{Rotating dToF camera}
Rotating dToF sensors, widely used in automotive LiDAR systems, acquire depth by sweeping laser beams across the scene, %
producing line-structured and view-dependent depth patterns. To simulate this sensing mechanism, we follow the specifications of the Velodyne VLP-16 and VLP-32 families. 
For each simulated scan, we sample ray directions stochastically within the parameter ranges spanned by these devices, including horizontal resolutions of 0.33\degree–2\degree, vertical resolutions of 0.1\degree–0.4\degree, vertical fields of view from $-30\degree$ to $30\degree$, and Gaussian noise levels up to a standard deviation of 0.001.
This stochastic configuration yields realistic variations in scan density, angular spacing, and measurement noise,
capturing the diverse sampling characteristics of real rotating dToF sensors.

\mparagraph{Noise augmentations}
In addition to modeling the characteristics of different dToF devices, we introduce a set of data augmentation strategies that simulate the degradation patterns commonly observed in real-world sensing conditions. 
dToF measurements frequently exhibit missing depth on highly reflective or low-return surfaces and under strong ambient illumination. 
Multipath interference around transparent or specular objects can produce erroneous depth values, while depth drift may occur due to object motion, sensor vibration, or imperfect calibration. Because these artifacts are highly device-dependent, we incorporate randomized perturbations to encourage our model to become device-agnostic.

Concretely, we add Gaussian noise proportional to depth magnitude, apply random spatial jitter, and inject 0.2--1.0\% outlier points whose depths are uniformly sampled between the near and far planes of the ground-truth depth map. 
To further mimic occlusions and unobserved regions, we apply depth inpainting augmentation, which randomly removes square patches, edge-aligned regions, and irregular masks generated using Perlin noise. 
These augmentations collectively expose the model to a broad spectrum of real-world degradation patterns, significantly improving robustness to diverse sensing environments.

\begin{table*}[th]
	\centering
	\small
	\setlength{\tabcolsep}{4pt}
    \caption{The comparison of zero-shot generalization across real and simulated sparse-depth inputs. Each dataset block reports the relative depth error (Rel) and threshold accuracy ($\delta$).
    	The best, second-best, and third-best results are highlighted in \cellcolor{darkgreen!50}dark, \cellcolor{darkgreen!30}medium, and \cellcolor{darkgreen!15}light green.} %
    \vspace{-2mm}
    \newcolumntype{C}{>{\centering\arraybackslash}m{0.6cm}}
    \label{tab:depth_completion}
    	\resizebox{0.85\textwidth}{!}{%
	\begin{tabular}{
            l| C C | C C | C C | C C | C C | C C | C C | r c
		}
		\thickhline
        Input & \multicolumn{4}{c|}{Real sensor depth} & \multicolumn{8}{c|}{Simulated points} & \multicolumn{4}{c}{}\\
        \hline
		\multirow{2}{*}{Method} &
		\multicolumn{2}{c|}{KITTI-DC} &
		\multicolumn{2}{c|}{ZJUL5} &
        \multicolumn{2}{c|}{DDAD} &
		\multicolumn{2}{c|}{DIODE} &
		\multicolumn{2}{c|}{ETH3D} &
		\multicolumn{2}{c|}{iBims-1} &
		\multicolumn{2}{c|}{Average} &
        \multicolumn{2}{c}{Inference Cost} \\
		\cline{2-17}
		& Rel$\downarrow$ & $\delta\uparrow$ & Rel$\downarrow$ & $\delta\uparrow$ & Rel$\downarrow$ & $\delta\uparrow$ & Rel$\downarrow$ & $\delta$$\uparrow$ & Rel$\downarrow$ & $\delta$$\uparrow$ & Rel$\downarrow$ & $\delta$$\uparrow$ & Rel$\downarrow$ & $\delta$$\uparrow$ & time $\downarrow$ & memory$\downarrow$ \\
		\hline
        OMNI-DC
        & \cellcolor{darkgreen!40} 1.48 & \cellcolor{darkgreen!40} 90.1
        & 12.92 & 29.8
        & \cellcolor{darkgreen!20} 5.15 & \cellcolor{darkgreen!40} 72.7
        & \cellcolor{darkgreen!40} 0.83 & \cellcolor{darkgreen!40}97.1
        & \cellcolor{darkgreen!10} 1.05 & \cellcolor{darkgreen!10}91.6
        & \cellcolor{darkgreen!40} 1.16 & \cellcolor{darkgreen!40} 92.5
        & \cellcolor{darkgreen!20} 3.77 & \cellcolor{darkgreen!40} 79.0 & 638 ms & 3.49 GB \\
        PromptDA\scriptsize(ViT-S) &  2.90 & 72.6 & 13.25 & 29.4 & 7.20 & 49.4 & 2.11 & 88.7 & 2.67 & 76.7 & 2.42 & 81.2 & 5.09 & 66.3 & \cellcolor{darkgreen!40} 30 ms & \cellcolor{darkgreen!40} 0.40 GB\\
        PromptDA\scriptsize(ViT-L) & 3.38 & 65.1 & 12.8 & 30.1 & 7.32 & 44.7 & 2.27 & 88.1 & 2.71 & 78.2 & 2.37 & 83.3 & 5.14 & 64.9 & \cellcolor{darkgreen!10} 102 ms & 1.56 GB\\ 
        PriorDA\scriptsize(ViT-S)
        & 2.85    & 75.2
        & \cellcolor{darkgreen!10} 11.53   & \cellcolor{darkgreen!10} 32.1
        & \cellcolor{darkgreen!10} 6.36    & 60.5
        & 1.27    & 92.9
        & 1.30    & 89.6
        & 1.52    & 88.7
        & 4.14    & 73.2
        & 118 ms & \cellcolor{darkgreen!10} 0.77 GB\\
        PriorDA\scriptsize(ViT-B)
        & \cellcolor{darkgreen!10} 2.76  & \cellcolor{darkgreen!10} 76.3
        & \cellcolor{darkgreen!40} 11.13 & \cellcolor{darkgreen!20} 32.2
        & 6.50  & \cellcolor{darkgreen!10} 60.9
        & \cellcolor{darkgreen!10} 1.15  & \cellcolor{darkgreen!10} 93.9
        & \cellcolor{darkgreen!20} 1.04  & \cellcolor{darkgreen!20} 92.9
        & \cellcolor{darkgreen!10} 1.35  & \cellcolor{darkgreen!10} 90.6
        & \cellcolor{darkgreen!10} 3.99  & \cellcolor{darkgreen!10} 74.5
        & 197 ms & 1.62 GB\\
        Marigold-DC & 5.50 & 43.8 & 12.77 & 29.4 & 12.01 & 29.8 & 2.58 & 82.3 & 1.80 & 83.4 & 2.00 & 85.8 & 6.11 & 59.1 & 6470 ms& 5.71 GB\\
		\hline
        Ours
        & \cellcolor{darkgreen!20}2.00  & \cellcolor{darkgreen!20} 84.0
        & \cellcolor{darkgreen!40}11.13 & \cellcolor{darkgreen!40} 32.3
        & \cellcolor{darkgreen!40}4.61 & \cellcolor{darkgreen!20} 68.8
        & \cellcolor{darkgreen!20}0.89 & \cellcolor{darkgreen!20} 96.7
        & \cellcolor{darkgreen!40} 0.84 & \cellcolor{darkgreen!40} 94.8
        & \cellcolor{darkgreen!20} 1.29 &  \cellcolor{darkgreen!20} 90.9
        & \cellcolor{darkgreen!40}3.46 & \cellcolor{darkgreen!20} 77.9
        & \cellcolor{darkgreen!20} 34 ms & \cellcolor{darkgreen!20} 0.44 GB \\
		\thickhline
	\end{tabular}
		}
\end{table*}

\begin{table*}[h]
	\centering
	\small
	\setlength{\tabcolsep}{4pt}
	\newcolumntype{C}{>{\centering\arraybackslash}m{0.7cm}}
    \vspace{-2mm}
    \caption{The comparison of zero-shot generalization under varying noise patterns: spatial shifting (\textit{spa.}), precision degradation (\textit{deg.}) and missing holes (\textit{mis.}). Each dataset block shows relative error (Rel) of metric depth in percentage.
    	The best, second-best, and third-best results are highlighted in \cellcolor{darkgreen!50}dark, \cellcolor{darkgreen!30}medium, and \cellcolor{darkgreen!15}light green, respectively.}
    \label{tab:noise_robustness}
    \vspace{-3mm}
	\resizebox{0.85\textwidth}{!}{%
	\begin{tabular}{
        l| C C C|C C C|C C C|C C C|C C C}
		\thickhline
		\multirow{2}{*}{Method} &
		\multicolumn{3}{|c}{DDAD} &
		\multicolumn{3}{|c}{DIODE} &
		\multicolumn{3}{|c}{ETH3D} &
		\multicolumn{3}{|c}{iBims-1} &
		\multicolumn{3}{|c}{Average} \\
		\cline{2-16}
		& \textit{spa}. & \textit{deg}. & \textit{mis.} 
        & \textit{spa}. & \textit{deg}. & \textit{mis.} 
        & \textit{spa}. & \textit{deg}. & \textit{mis.} 
        & \textit{spa}. & \textit{deg}. & \textit{mis.}  
        & \textit{spa}. & \textit{deg}. & \textit{mis.}  \\
		\hline
        OMNI-DC     & \cellcolor{darkgreen!20} 6.84 & \cellcolor{darkgreen!20}  5.28 & \cellcolor{darkgreen!20} 6.66 & \cellcolor{darkgreen!20} 1.16 & \cellcolor{darkgreen!20} 1.24 & \cellcolor{darkgreen!20} 1.09 & \cellcolor{darkgreen!20} 1.60 & \cellcolor{darkgreen!20} 1.53 & \cellcolor{darkgreen!10} 1.62 & \cellcolor{darkgreen!20} 1.54 & \cellcolor{darkgreen!20} 1.67 & \cellcolor{darkgreen!40} 1.35 & \cellcolor{darkgreen!20} 2.78 & \cellcolor{darkgreen!20} 2.43 & \cellcolor{darkgreen!20} 2.68 \\
		PromptDA    & 8.50 & 7.27 & 10.1 & 2.21 & 2.23 & 2.56 & 2.87 & 2.85 & 3.41 & 2.50 & 2.62 & 2.88 & 4.02& 3.74& 4.73 \\
		PriorDA     & \cellcolor{darkgreen!10} 7.26 & \cellcolor{darkgreen!10} 6.41 & \cellcolor{darkgreen!10} 7.10 & \cellcolor{darkgreen!10} 1.42 & \cellcolor{darkgreen!10} 1.44 & \cellcolor{darkgreen!10} 1.40 & \cellcolor{darkgreen!10} 1.66 & \cellcolor{darkgreen!10} 1.56 & \cellcolor{darkgreen!20} 1.47 & \cellcolor{darkgreen!10} 1.71 & \cellcolor{darkgreen!10} 1.75 & \cellcolor{darkgreen!10} 1.63 & \cellcolor{darkgreen!10} 3.01 & \cellcolor{darkgreen!10} 2.79 & \cellcolor{darkgreen!10} 2.90 \\
		Marigold-DC & 13.70& 12.05& 13.34& 3.16& 2.76& 2.66& 2.25& 2.04& 1.90 & 2.49 & 2.57 & 2.08  & 5.40 & 4.97 &  4.99  \\ 
		\hline
		Ours    & \cellcolor{darkgreen!40} 5.11 & \cellcolor{darkgreen!40} 4.64 & \cellcolor{darkgreen!40}5.37 & \cellcolor{darkgreen!40} 0.97 & \cellcolor{darkgreen!40} 0.97 & \cellcolor{darkgreen!40} 1.04 & \cellcolor{darkgreen!40} 1.01 & \cellcolor{darkgreen!40} 1.01 & \cellcolor{darkgreen!40} 1.02 & \cellcolor{darkgreen!40} 1.39 & \cellcolor{darkgreen!40} 1.48 & \cellcolor{darkgreen!20} 1.48 & \cellcolor{darkgreen!40} 2.12 & \cellcolor{darkgreen!40} 2.02 & \cellcolor{darkgreen!40} 2.23 \\
		\thickhline
	\end{tabular}
    	}%
    	\vspace{-1mm}
\end{table*}

\begin{table}[h]
\vspace{-0.3cm}
	\centering
	\small
    \caption{Ablation study on the network architecture.
    	We evaluate against decoder-level depth prompting following PromptDA, and joint attention without a directional masking.
    We also scale the network backbone to evaluate performance across model capacities.
    We report the average relative depth error (Rel) and threshold accuracy ($\delta$). Refer to supplements for the full table.}
    \vspace{-0.2cm}        
    \newcolumntype{C}{>{\centering\arraybackslash}m{1.2 cm}}
    \label{tab:ablation_arch}
    \resizebox{0.85\linewidth}{!}{%
    \begin{tabular}{l|  C C}
		\thickhline
		Methods& Rel & $\delta$ \\
		\hline
        Depth prompting & 4.07 & 73.6 \\
        Joint attention & 3.87 & 75.6 \\
        Masked joint attention (Ours, ViT-S) & 3.46 & 77.9 \\
        \hline
        Ours, ViT-B & 3.28 & 79.0 \\
        Ours, ViT-L & 3.07 & 80.1 \\
		\thickhline
	\end{tabular}
}
     \vspace{-5mm}
\end{table}

\begin{figure}[t]
	\centering
	\vspace{-2mm}
	\resizebox*{0.95 \linewidth}{!}{
		\includegraphics[width=\linewidth]{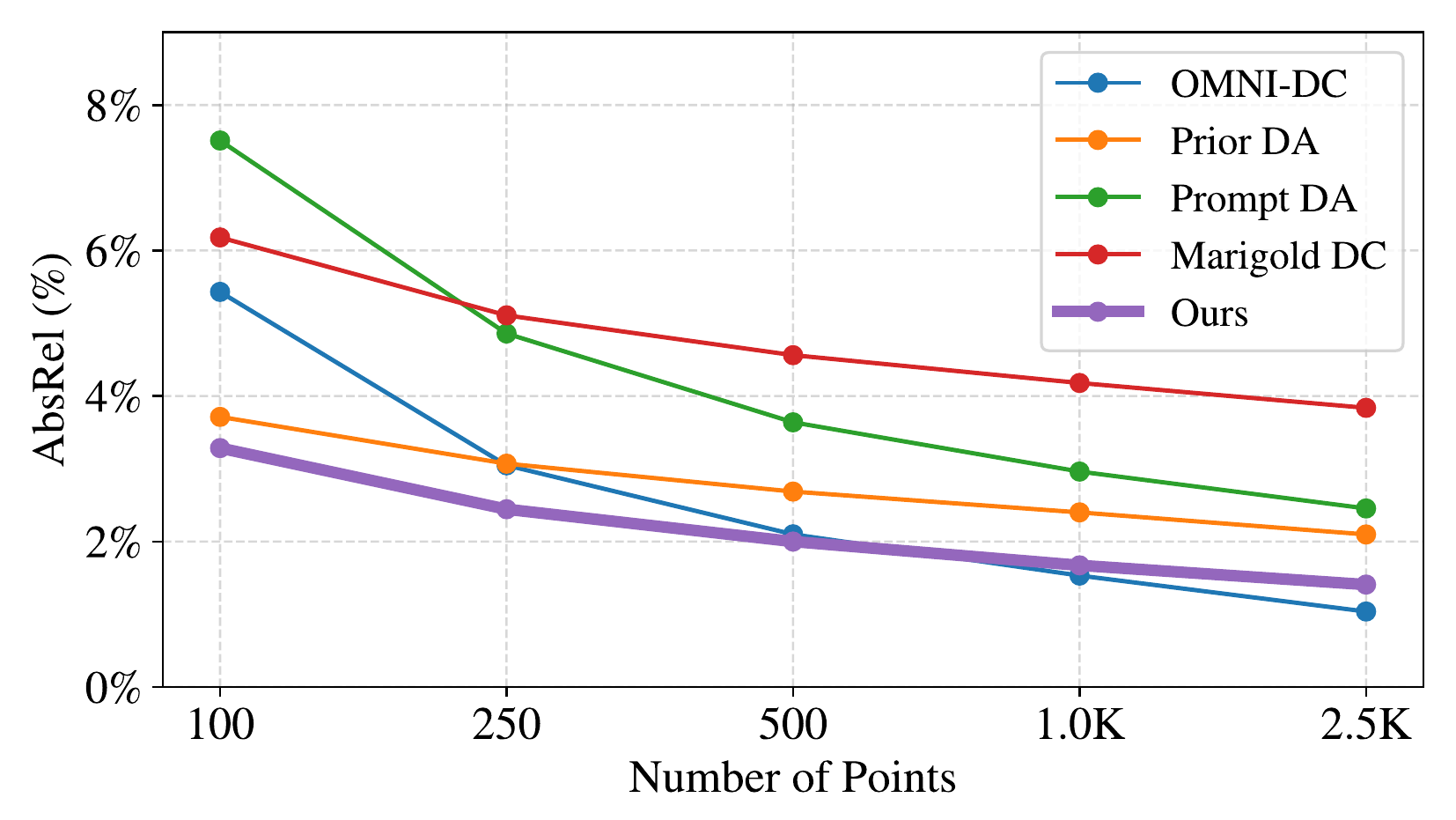}
	}
	\vspace{-4mm}
	\caption{Quantitative analysis on varying number of depth inputs. Refer to supplements for numerical details.}
	\vspace{-7mm}
	\label{fig:density_analysis}
\end{figure}

\section{Experiments}
\subsection{Implementation Details}
During preprocessing, we apply a flood-fill algorithm to complete missing-depth regions before constructing the unified depth representation. Our encoder is initialized with ViT-Small weights pretrained using DINOv2, and we extract features from the 6th and 12th intermediate layers to feed into the DPT-style decoder. Following MoGe-2~\cite{wang2025moge2}, the decoder adopts a multi-scale architecture, but we reduce the feature dimensions of the residual blocks to 384, 256, 64, 32, and 16 for improved efficiency. Our implementation also supports a dynamic number of depth tokens; during training, the number of RGB tokens is sampled from $[800, 2000]$, while the number of depth tokens varies between $[800, 1200]$ to reflect the input sparsity.
We train the model using the AdamW optimizer with learning rates of $1\times10^{-5}, 1\times10^{-4}$ for the encoder and the decoder, respectively, halving the learning rate every 25k iterations. During the first 50k iterations, we adopt a low-resolution warm-up stage to stabilize training. For all evaluations, the final model is trained for 100k iterations on 8 NVIDIA A100 GPUs.

\subsection{Evaluation Setting}
We evaluate our approach under a strict zero-shot generalization setting using both real-sensor and simulated benchmarks.
For real-sensor evaluation, we use KITTI-DC~\cite{Uhrig2017THREEDV}, captured with a Velodyne LiDAR, and ZJUL5~\cite{deltar}, obtained from the lightweight VL53L5CX sensor. These datasets present extremely sparse inputs (down to $8 \times 8$), span both indoor and outdoor scenes, and represent two distinct sensor families.

For simulated benchmarks, we randomly sample 500 sparse depth points from the ground-truth depth maps.
We use DDAD~\cite{packnet} for outdoor driving scenes, DIODE~\cite{diode_dataset} and ETH3D~\cite{schoeps2017cvpr} for mixed indoor–outdoor scenarios, and iBims-1~\cite{koch2018evaluation} for indoor evaluations.
As these datasets provide real sensor measurements, we follow the common depth completion protocol that generates sparser inputs via downsampling. Notably, DDAD is captured using Luminar-H2, adding another real dToF sensor under evaluation.

We compare against state-of-the-art depth completion and enhancement approaches, including OMNI-DC~\cite{zuo2025omni}, Marigold-DC~\cite{viola2025marigold}, PromptDA~\cite{lin2025prompting}, and Prior Depth Anything~\cite{wang2025depth}. As PromptDA does not natively support sparse depth,
we complete missing regions with a flood-fill operation after nearest-neighbor downsampling to match a ViT-compatible resolution.
Performance is reported using the relative error \( \mathrm{rel} = \| \tilde{\mathbf{d}} - \mathbf{d} \|_1 / \mathbf{d} \) and the threshold accuracy \( \delta = \max\!\left( \tilde{\mathbf{d}} / {\mathbf{d}}, \, \mathbf{d} / \tilde{\mathbf{d}} \right) < 1.025 \).

\begin{figure*}[ht]
	\centering
	\vspace{-2mm}
	\adjustbox{max width=\linewidth}{
		\includegraphics[width=\linewidth]{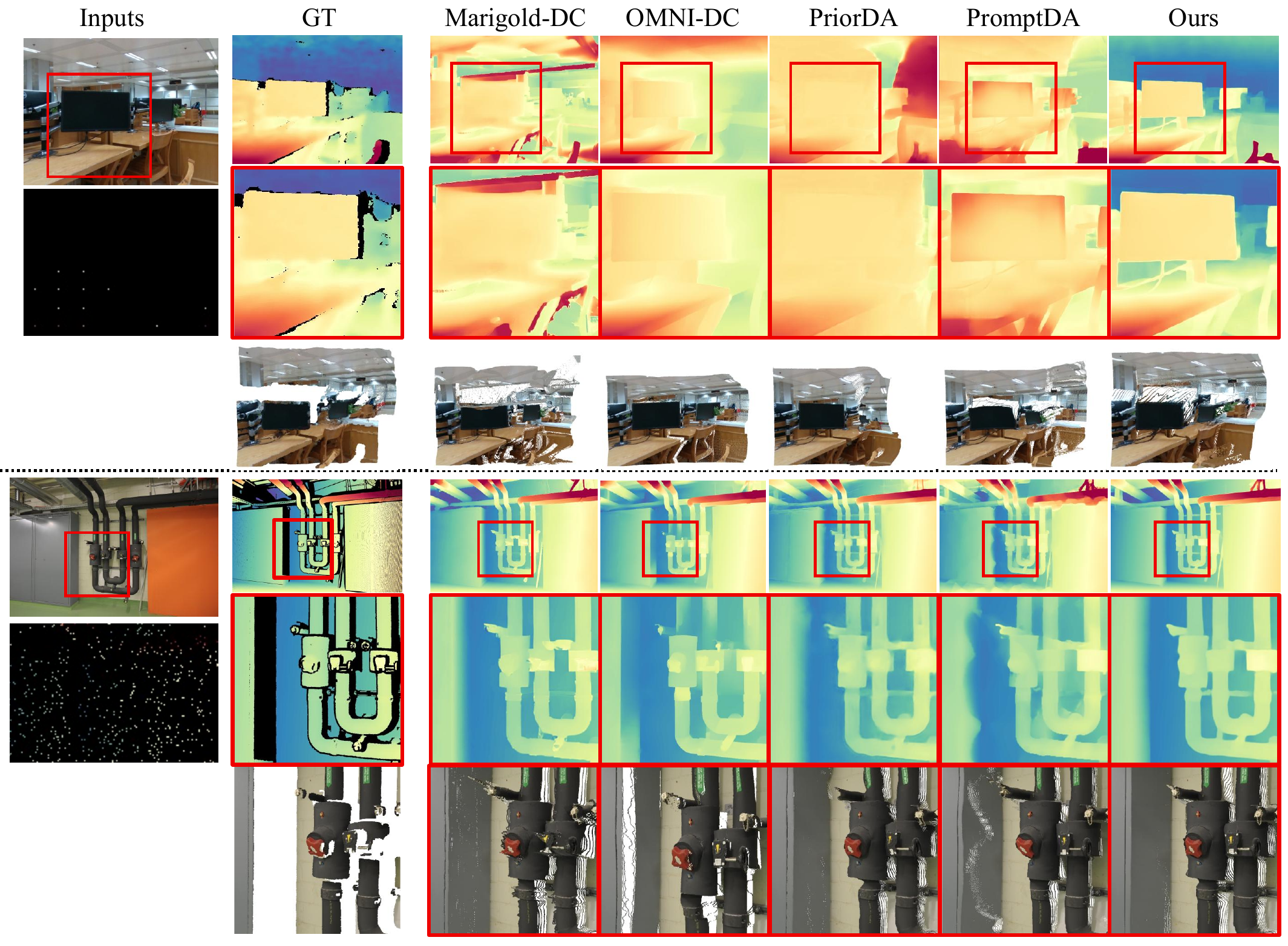}
	}
	\vspace{-8mm}
	\caption{Qualitative comparison of metric depth maps, and reconstructed point clouds. The two examples are taken from the Library scene of ZJUL5, and the Pipes scenes of ETH3D.}
    \vspace{-5mm}
	\label{fig:main_result}
\end{figure*}

\subsection{Quantitative Evaluation}
\mparagraph{Depth completion}
Table~\ref{tab:depth_completion} shows that our method achieves competitive or state-of-the-art results across all benchmarks while maintaining high computational efficiency. Notably, our model achieves the best performance on ETH3D and ZJUL5. ETH3D includes high-resolution DSLR imagery ($2048 \times 1365$), while ZJUL5 provides extremely low-resolution dToF measurements. These results highlight our model’s ability to perform accurate metric depth reconstruction even when the input depth is vastly sparser than the image resolution.

We also compare inference speed and memory usage on an NVIDIA A100 GPU. PromptDA and our model are evaluated in FP16, while other methods follow their official settings.
All timings are measured with batch size 1 and averaged across benchmarks at native input resolutions. %
Please refer to the supplementary material for the benchmark resolution specifications.
Compared to OMNI-DC, our model runs 20× faster and requires 10× less memory, while achieving comparable or higher accuracy—demonstrating the strong efficiency–accuracy trade-off offered by our approach.

\mparagraph{Robustness to noise}
We further assess robustness to three common depth degradation types: spatial shifting, precision degradation, and missing-depth holes. Experiments are conducted on DDAD, DIODE, ETH3D, and iBims-1: 
(1) Spatial shifting: 500 points are randomly sampled and perturbed by spatial jitter up to $0.5\%$ of the image diagonal.
(2) Precision degradation: we introduce edge distortions via random erosion/dilation of the ground-truth depth, with erosion thickness set to 1\% of the image diagonal, followed by sampling 500 points.
(3) Missing holes: we remove depth regions using square masks or irregular Perlin-noise masks covering $20 \pm 5\%$ of the image.

We evaluate using relative error, consistent with the main experiments. As shown in Table~\ref{tab:noise_robustness}, our method achieves the best results across all noise types%
, indicating strong resilience to depth acquisition artifacts.

\mparagraph{Robustness to depth sparsity}
To analyze performance under varying sparsity, we sample 100–2.5k points from ground-truth depth of DDAD, DIODE, ETH3D, and iBims-1. Relative errors are averaged across datasets and visualized in Figure~\ref{fig:density_analysis}. Our method achieves the highest accuracy in the extreme sparse regime (100 points) and exhibits the smallest performance drop as the number of points increases, demonstrating strong robustness to input density variations.

\subsection{Qualitative Comparison}
Figure~\ref{fig:main_result} presents qualitative comparisons. For each method, we visualize the predicted metric depth, and the reconstructed 3D point cloud. Our method consistently recovers more accurate global geometry, particularly in regions with extremely sparse or missing depth.
In the Library scene (row 1), competing methods significantly misestimate the ceiling distance, while our model reconstructs the correct global structure.
In the Pipes scene (row 2), competing methods estimate distorted pipes, whereas our approach preserves a correct geometry. Our approach preserves correct spatial relationships of local geometry and even for far-range structures with few depth observations.

\subsection{Ablation study}
\mparagraph{Depth fusion architecture}
We validate the importance of masked joint attention through two baselines:
(1) a depth-prompting variant that follows PromptDA by injecting depth features only into the decoder’s convolutional neck, and
(2) joint attention without directional masking.
All variants are trained under identical settings. As shown in Table~\ref{tab:ablation_arch}, encoder-level fusion already outperforms decoder-level prompting, and introducing the directional mask further improves accuracy. These results confirm that masked joint attention is a crucial architectural component.

\mparagraph{ViT backbone}
We also report the results of our model using a ViT-Base and a ViT-Large backbone in Table~\ref{tab:ablation_arch}. The performance further improves with the larger ViT model, demonstrating the scalability of our model. 
Note, our ViT-Small model already outperforms PromptDA and PriorDA that rely on larger ViT-B or ViT-L backbones.
Our method is model size-agnostic by design, and we focus on ViT-Small to emphasize efficiency and practical deployment.

\section{Conclusion}
We presented a lightweight and generalizable framework for dense metric depth completion from sparse dToF measurements. Our depth-guided dual-branch encoder enables effective RGB and sparse depth fusion, while our dToF simulation pipeline greatly improves robustness across devices, sparsity levels, and noise conditions. Experiments on real and simulated benchmarks show that our method achieves strong accuracy and efficiency, establishing a practical solution for dense depth estimation from sparse dToF inputs.

\appendix
\vspace{-0.1cm}
\section*{Acknowledgements}
\vspace{-0.15cm}
\noindent Min H.~Kim acknowledges the Samsung Research Funding \& Incubation Center (SRFC-IT2402-02), the Korea NRF grant (RS-2024-00357548), the MSIT/IITP of Korea (RS-2022-00155620, RS-2024-00398830, RS-2024-00436680, and 2017-0-00072), the MSIT Advanced GPU Utilization Support Program, and Microsoft Research Asia.

\clearpage

{
    \small
    \bibliographystyle{ieeenat_fullname}
    \bibliography{dtof_dense_depth}
}

\clearpage 
\pagestyle{empty} 


\section*{Supplementary material}

# Dense Metric Depth Completion from Sparse Direct Time-of-Flight Sensors

## Supplementary Material

Table I. List of train datasets with domain, frame count, and sampling weight.

| Name | Domain | #Frames | Weight |
|---|---|---|---|
| ApolloSynthetic | Outdoor/driving | 194k | 3.8 |
| EDEN | Outdoor/garden | 369k | 1.2 |
| GTA-SfM | Outdoor/in-the-wild | 19k | 2.7 |
| Hypersim | Indoor | 75k | 4.8 |
| IRS | Indoor | 91k | 5.4 |
| KenBurns | In-the-wild | 76k | 1.5 |
| MatrixCity | Outdoor/driving | 391k | 1.3 |
| MidAir | Outdoor/in-the-wild | 424k | 3.8 |
| MVS-Synth | Outdoor/driving | 12k | 1.2 |
| ObjaverseV1 | Object | 168k | 4.6 |
| Structured3D | Indoor | 77k | 4.6 |
| Synscapes | Outdoor/driving | 25k | 1.9 |
| Synthia | Outdoor/driving | 96k | 1.1 |
| UrbanSyn | Outdoor/driving | 7.5k | 2.0 |
| UnrealStereo4K | In-the-wild | 8k | 1.6 |
| TartanAir | In-the-wild | 293k | 4.8 |

## 1. Additional Results

**Affine depth alignment comparison.** We compare our method against state-of-the-art RGB-only MDE models with post-hoc sensor depth alignment to assess whether sparse dToF can be replaced by scale alignment alone. We align Depth Anything v2 and MoGe-1 predictions to sparse sensor depth using least-squares fitting, and additionally report robust ROE alignment for MoGe. As shown in Table IV, our method consistently outperforms these aligned baselines, demonstrating that encoder-stage fusion provides substantially richer geometric guidance than post-hoc alignment, even with strong monocular priors.

**Full table of performance on varying input depth sparsity.** For clarity, we report the full quantitative results in Tab. VII on varying input depth sparsity, which corresponds to the plot shown in Figure 6 of the main paper. The result shows that our method achieves best score in extreme sparse inputs and exhibits a smaller performance drop as sparsity increases compared to all competing methods, except for PriorDA, which has lower accuracy. Both the visualization and the table indicate that our model achieves superior relative stability under variations in input depth sparsity.

**Depth completion qualitative results.** We present qualitative comparisons for depth completion across all evaluation benchmarks, using both metric depth maps and point cloud visualizations. Figure I shows examples from ETH3D and iBims-1, where our method, especially in the Electro scene (row 1), accurately recovers thin structures in areas with sparse depth measurements. Results on DDAD in Figure II further demonstrate that our model correctly reconstructs trees (red box, row 1) and a telegraph pole (red arrow, row 2). Figure III presents qualitative results on DIODE. While existing methods fail to preserve local geometry, such as the metal bottle (row 1) or distort the lecture-room floor (row 2), our method maintains these structures and their spatial consistency. Finally, Figure IV shows results on real sensor datasets, including KITTI-DC and ZJUL5. Extremely sparse depth and textureless regions cause state-of-the-art methods fail to distinguish between a thin ornament with the bookshelf (row 1) or collapse desk legs into the foreground (row 2), whereas our method avoids such artifacts. In outdoor scenes such as KITTI-DC, our estimation prevents background regions from leaking into foreground objects.

Table II. Ablation study on flood-fill preprocessing and loss combinations. We report the average relative depth error (Rel) and threshold accuracy ($\delta$) over evaluation benchmarks: KITTI-DC, ZJUL5, DDAD, DIODE, ETH3D, and iBims-1.

| Methods | Rel | $\delta$ |
|---|---|---|
| Ours, w/o flood-fill | 3.54 | 76.7 |
| $\mathcal{L}_1 + \mathcal{L}_m$ | 3.53 | 77.4 |
| $\mathcal{L}_1 + \mathcal{L}_g + \mathcal{L}_m$ | 3.47 | 77.8 |
| $\mathcal{L}_1 + \mathcal{L}_g + \mathcal{L}_l + \mathcal{L}_m$ (Ours) | 3.46 | 77.9 |

Table III. Ablation study on number of the depth tokens $N_d$. We report the average relative depth error (Rel) and threshold accuracy ($\delta$) over evaluation benchmarks: KITTI-DC, ZJUL5, DDAD, DIODE, ETH3D, and iBims-1. We measure inference speed under the same evaluation settings as the main paper.

| Method | | Inference time | Rel | $\delta$ |
|---|---|---|---|---|
| | $N_d = 400$ | 32 ms | 3.87 | 71.9 |
| **Ours** | $N_d = 800$ | 32 ms | 3.55 | 76.5 |
| | $N_d = 1000$ | 34 ms | 3.47 | 77.4 |
| | $N_d = 1200$ | 34 ms | 3.47 | 77.6 |
| | $N_d = 2000$ | 36 ms | 3.43 | 78.1 |

## 2. Dataset details

**Evaluation datasets.** We utilize all images in the ETH3D (454 frames), iBims-1 (100 frames), and ZJUL5 datasets (1021 frames) for evaluation. For DDAD, we randomly select 1000 frames from the validation set. For KITTI-DC and DIODE, we use official validation splits. To ensure reliable evaluation, we further remove boundary artifacts in DIODE by detecting and masking out edge regions. For clarity, we additionally provide the input resolution used when reporting inference time. The resolutions are: KITTI-DC ($1216 \times 352$), ZJUL5 ($640 \times 480$), DDAD ($1936 \times 973$), DIODE ($1024 \times 768$), ETH3D ($2048 \times 1365$), and iBims-1 ($2048 \times 1365$).

**Training dataset.** Table I lists the synthetic datasets used to train our model. Our synthetic dataset covers diverse domains, including both indoor and outdoor scenes, as well as a wide variety of objects. The sampling ratio for training

Table IV. Comparison of metric depth accuracy after affine alignment of RGB-only monocular depth estimation models to sensor depth. Each dataset block reports the relative depth error (Rel) and threshold accuracy ($\delta$). Best and second-best results are **bold** and <u>underlined</u>.

| Method | KITTI-DC | | ZJUL5 | | DDAD | | DIODE | | ETH3D | | iBims-1 | | Average | |
|---|---|---|---|---|---|---|---|---|---|---|---|---|---|---|
| | Rel↓ | $\delta$ ↑ | Rel↓ | $\delta$ ↑ | Rel↓ | $\delta$ ↑ | Rel↓ | $\delta$ ↑ | Rel↓ | $\delta$ ↑ | Rel↓ | $\delta$ ↑ | Rel↓ | $\delta$ ↑ |
| DAv2+LS | 9.87 | 27.7 | 13.19 | 25.3 | 60.76 | 26.1 | 16.39 | 49.8 | 6.32 | 47.9 | 3.96 | 59.2 | 18.41 | 39.4 |
| MoGe+LS | 9.19 | 19.5 | 11.39 | 26.4 | 12.93 | 20.0 | 6.02 | 58.3 | 3.49 | 60.8 | 3.07 | 70.2 | 7.68 | 42.5 |
| MoGe+ROE | <u>4.61</u> | <u>48.7</u> | **10.30** | **34.2** | <u>8.46</u> | <u>37.0</u> | <u>3.05</u> | <u>75.7</u> | <u>2.67</u> | <u>71.6</u> | <u>2.50</u> | <u>77.4</u> | <u>5.27</u> | <u>57.5</u> |
| Ours (ViT-S) | **2.00** | **84.0** | <u>11.13</u> | <u>32.3</u> | **4.61** | **68.8** | **0.89** | **96.7** | **0.84** | **94.8** | **1.29** | **90.9** | **3.46** | **77.9** |

Table V. Full table of ablation study on the depth fusion architecture. We compared the depth-prompting variant decoder, the joint attention encoder, and the masked joint attention encoder architecture on KITTI-DC, ZJUL5, DDAD, DIODE, ETH3D, and iBims-1 datasets. Each dataset block reports the relative depth error (Rel) and threshold accuracy ($\delta$). Best results in each group are **bolded**

| Method | KITTI-DC | | ZJUL5 | | DDAD | | DIODE | | ETH3D | | iBims-1 | | Average | |
|---|---|---|---|---|---|---|---|---|---|---|---|---|---|---|
| | Rel | $\delta$ | Rel | $\delta$ | Rel | $\delta$ | Rel | $\delta$ | Rel | $\delta$ | Rel | $\delta$ | Rel | $\delta$ |
| Depth prompting | 2.14 | 82.1 | 12.16 | 31.2 | 5.93 | 58.2 | 1.23 | 94.1 | 1.39 | 88.2 | 1.60 | 87.6 | 4.07 | 73.6 |
| Joint attention | 2.14 | 81.6 | 12.42 | 30.8 | 5.02 | 64.2 | 1.04 | 95.7 | 1.07 | 92.6 | 1.55 | 88.6 | 3.87 | 75.6 |
| Masked joint attention (Ours) | **2.00** | **84.0** | **11.13** | **32.3** | **4.61** | **68.8** | **0.89** | **96.7** | **0.84** | **94.8** | **1.29** | **90.9** | **3.46** | **77.9** |

Table VI. Full table of ablation study on the ViT backbone scale. We evaluated our method with ViT-S, ViT-B, and ViT-L model as the encoder backbone on KITTI-DC, ZJUL5, DDAD, DIODE, ETH3D, and iBims-1 datasets. Each dataset block reports the relative depth error (Rel) and threshold accuracy ($\delta$). Best results in each group are **bolded**

| Method | KITTI-DC | | ZJUL5 | | DDAD | | DIODE | | ETH3D | | iBims-1 | | Average | |
|---|---|---|---|---|---|---|---|---|---|---|---|---|---|---|
| | Rel | $\delta$ | Rel | $\delta$ | Rel | $\delta$ | Rel | $\delta$ | Rel | $\delta$ | Rel | $\delta$ | Rel | $\delta$ |
| Ours, ViT-S | 2.00 | 84.0 | 11.13 | 32.3 | 4.61 | 68.8 | 0.89 | 96.7 | 0.84 | 94.8 | 1.29 | 90.9 | 3.46 | 77.9 |
| Ours, ViT-B | 1.93 | 84.7 | 10.56 | 33.5 | 4.53 | 70.6 | 0.80 | 97.1 | 0.72 | 95.9 | 1.16 | 92.0 | 3.28 | 79.0 |
| Ours, ViT-L | **1.90** | **85.0** | **9.95** | **34.7** | **4.27** | **72.7** | **0.68** | **97.7** | **0.58** | **97.0** | **1.04** | **93.2** | **3.07** | **80.1** |

Table VII. Full table of quantitative results showing performance with varying numbers of depth input points. Each block reports the average relative depth error (%) over the evaluation benchmarks. Best and second-best results are **bold** and <u>underlined</u>.

| Method | # Depth Points | | | | |
|---|---|---|---|---|---|
| | 100 | 250 | 500 | 1000 | 2500 |
| OMNI-DC | 5.43 | <u>3.05</u> | 2.10 | **1.53** | **1.04** |
| Prior DA | <u>3.71</u> | 3.07 | 2.69 | 2.40 | 2.10 |
| Prompt DA | 7.51 | 4.86 | 3.64 | 2.96 | 2.45 |
| Marigold DC | 6.18 | 5.11 | 4.56 | 4.18 | 3.84 |
| Ours | **3.28** | **2.44** | **2.00** | <u>1.67</u> | <u>1.41</u> |

follows the weighting strategy of MoGe [1].

## 3. More ablation results

**Flood-fill.** We use flood-fill to convert sparse depth into a 2D representation for ViT tokenization, while a validity mask preserves sparsity and avoids discontinuity bias. Flood-fill introduces no artifacts and stabilizes learning under extreme sparsity, with density-invariant efficiency fixed by the 2D token budget. An ablation study in Table II confirms its effectiveness, as removing flood-fill degrades performance.

**Loss.** We conduct an ablation study on different loss combinations and evaluate the performance on the same datasets used in the depth completion experiments. The average relative depth error and threshold accuracy are used as evaluation metrics. As shown in Table II, incorporating the global scale-invariant loss $\mathcal{L}_g$ yields a slight performance gain, and further adding the local scale-invariant loss $\mathcal{L}_l$ provides an additional improvement.

**Number of depth tokens.** We also study the effect of varying the number of depth tokens at inference time. We evaluate models with 400–2000 depth tokens (trained with 800–1200). Table III shows that reducing depth tokens can decrease inference time by $\sim$4 ms with only a 0.4% increase in error, demonstrating a flexible trade-off between efficiency and accuracy.

**Full table of depth fusion architecture ablation.** We report detailed ablation results of the network architecture on each benchmark in Table V. The table shows that an encoder with masked joint attention consistently outperforms both the encoder with joint attention and the decoder based on depth prompting.

**Full table of ViT backbone ablation.** We include detailed results of ViT backbone ablation in Table VI. All variants are trained under identical settings for 100K iterations. The results show that increasing network scale consistently improves performance across all benchmarks, demonstrating that our proposed fusion architecture scales well with backbone capacity.

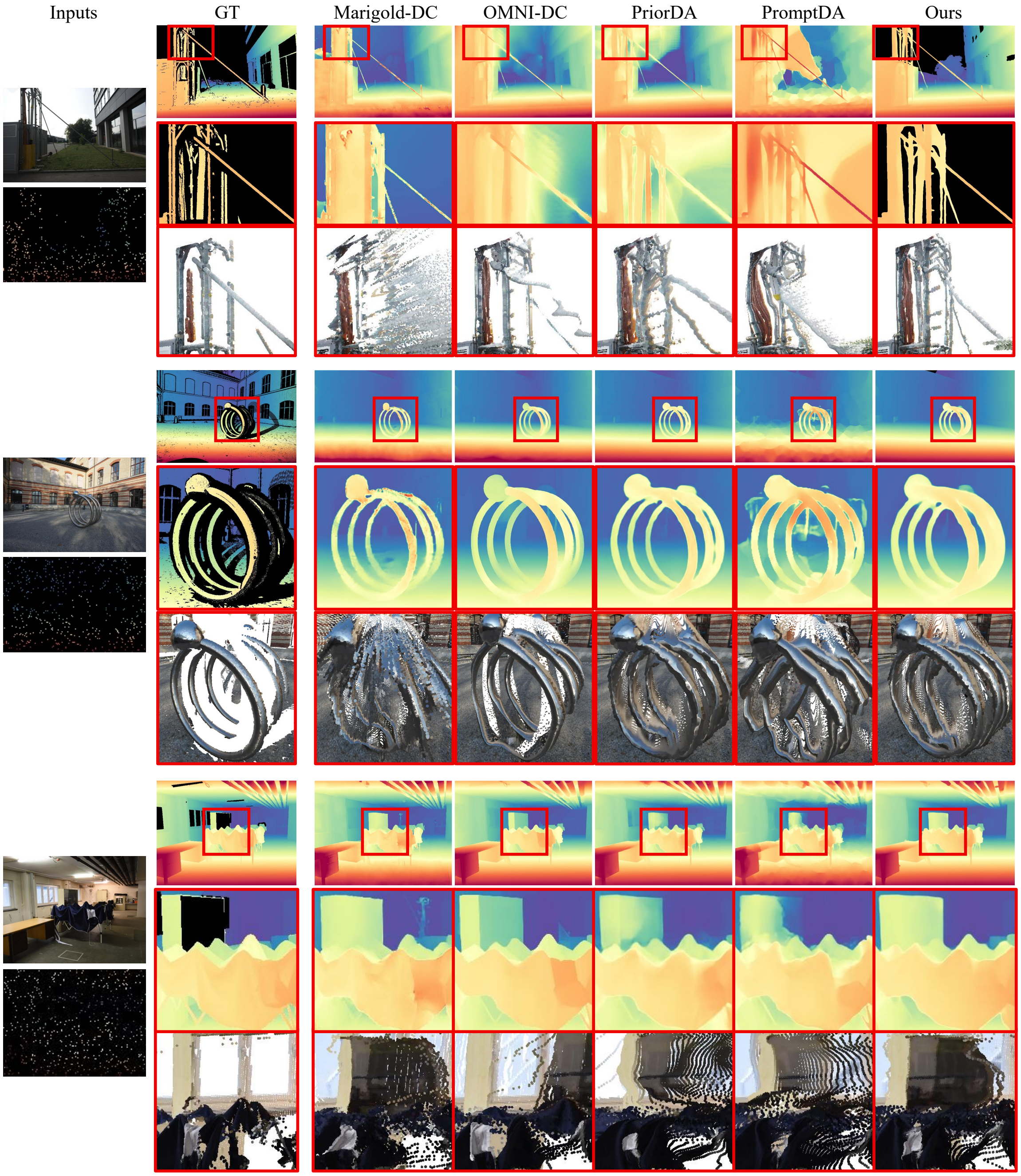


Figure I. Qualitative comparison of metric depth maps, and reconstructed point clouds. Examples are from the ETH3D dataset's electro (top), facade (middle) scenes, and iBims-1's factory (bottom) scene.

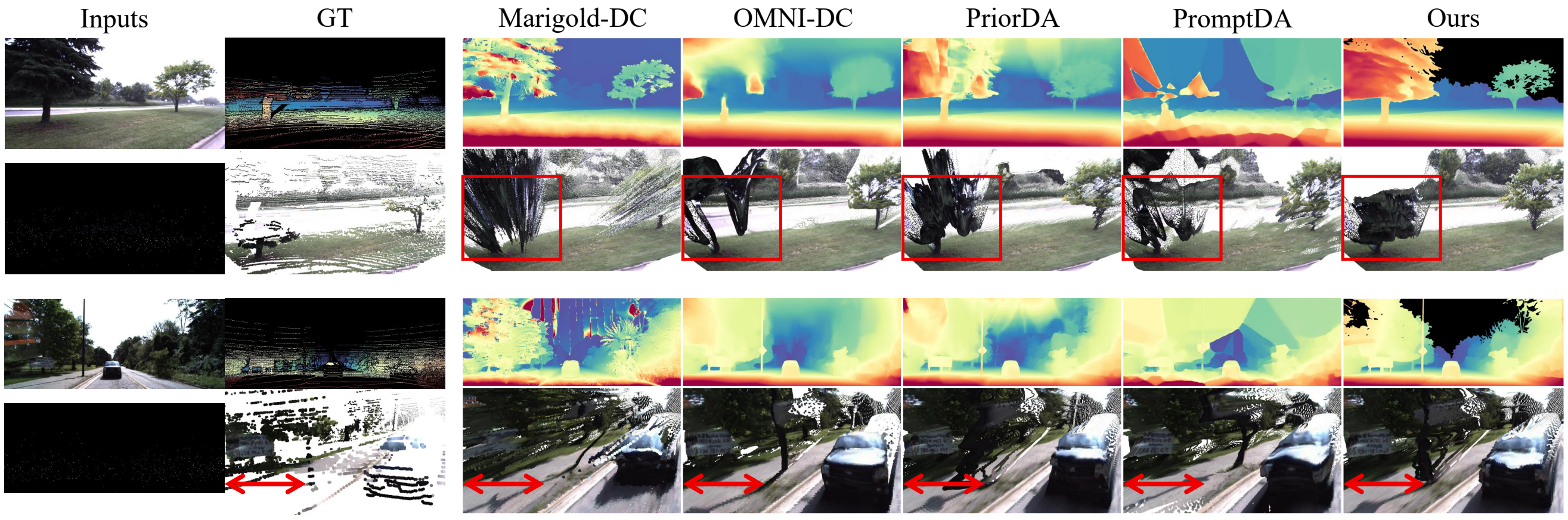


Figure II. Qualitative comparison of metric depth maps, and reconstructed point clouds. Examples are from DDAD dataset

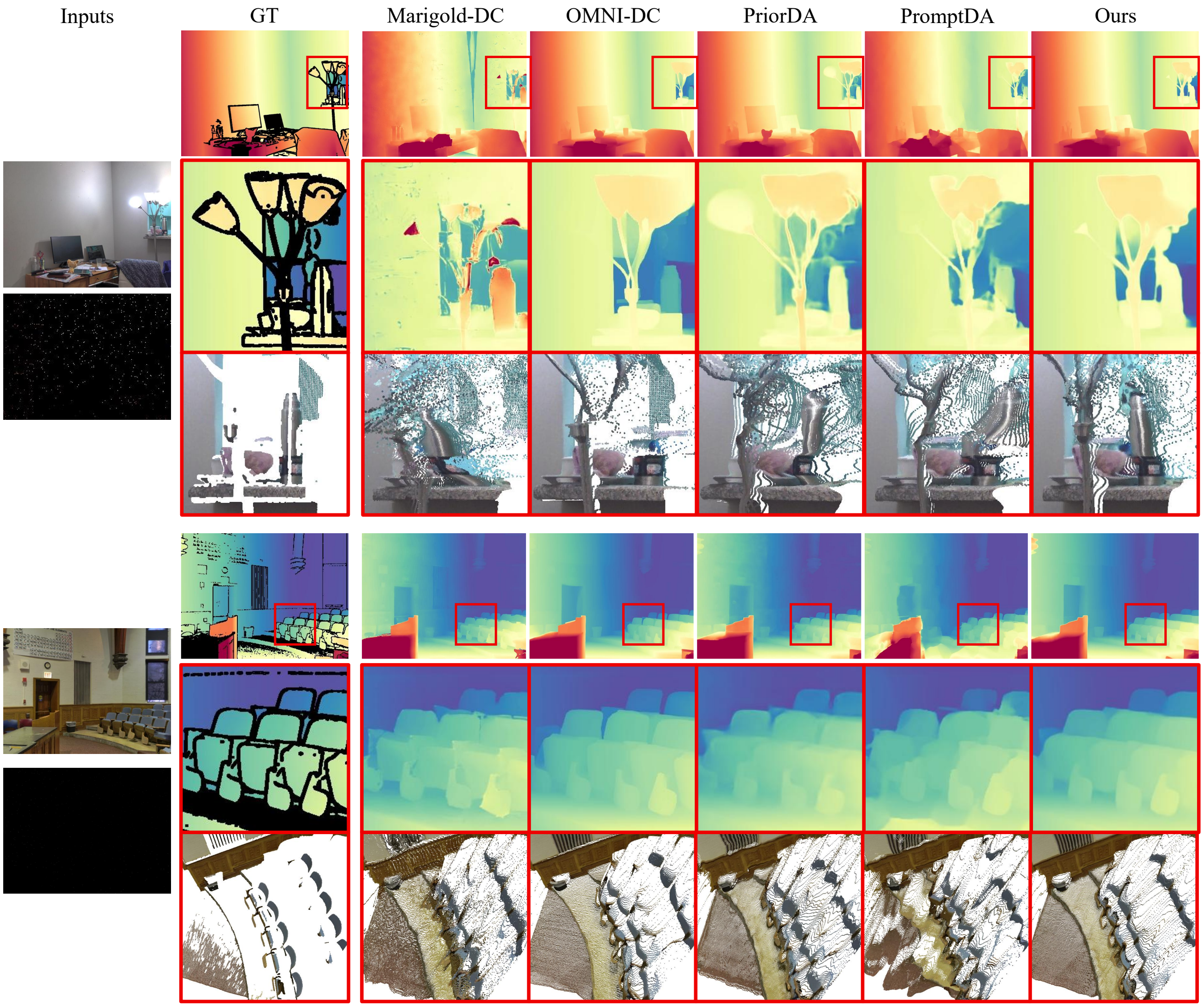


Figure III. Qualitative comparison of metric depth maps, and reconstructed point clouds. Examples are from the DIODE dataset

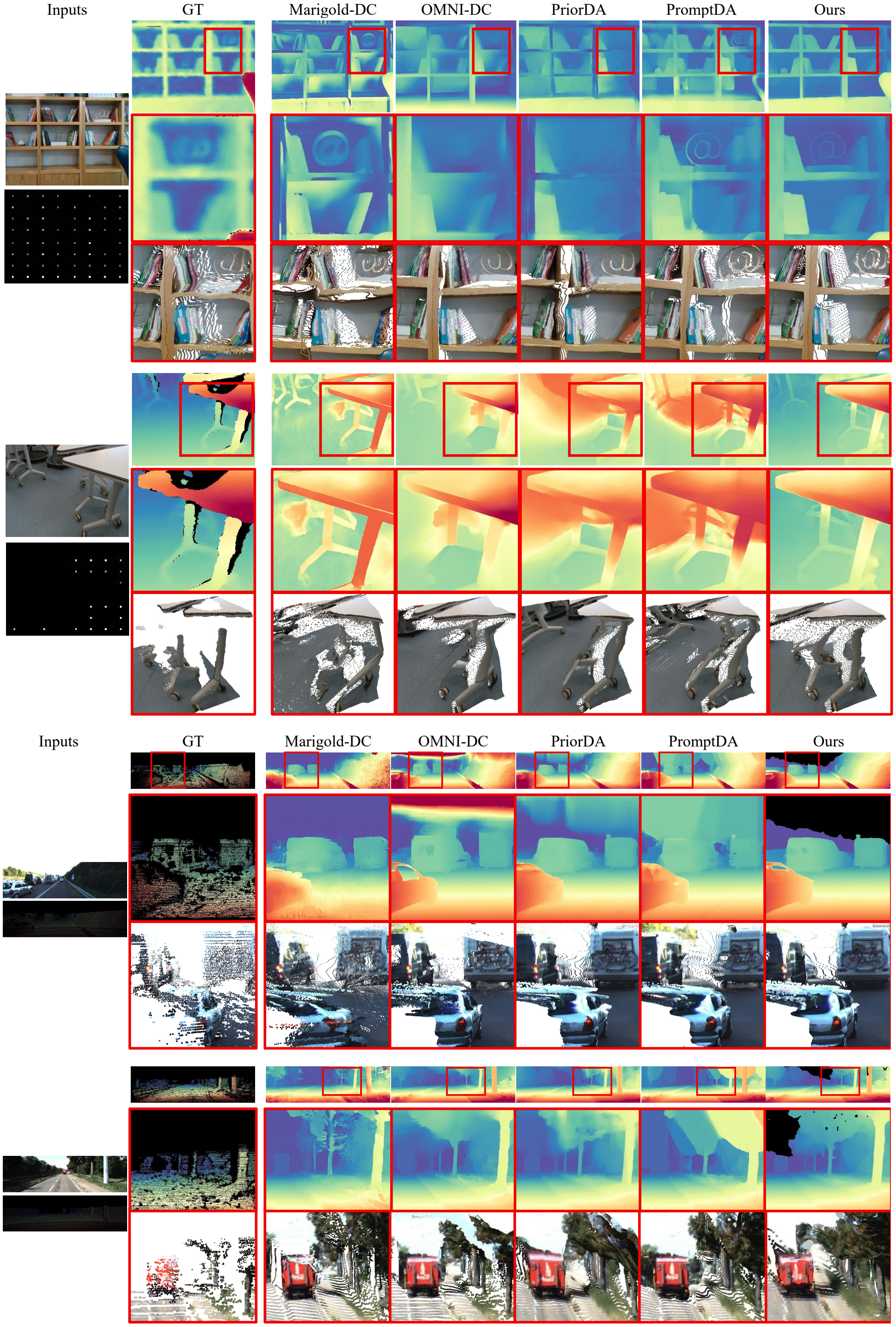


Figure IV. Qualitative comparison of metric depth maps, and reconstructed point clouds. The first row shows examples from the cafe1 scene and the second from the teaching region scene of the ZJUL5 dataset, while the third and fourth rows present examples from the KITTI-DC dataset.


\clearpage 
\pagestyle{plain} 

\end{document}